\documentclass[letterpaper]{article} 
\usepackage[]{aaai23}  
\usepackage{times}  
\usepackage{helvet}  
\usepackage{courier}  
\usepackage[hyphens]{url}  
\usepackage{graphicx} 
\usepackage{natbib}  
\usepackage{caption} 
\usepackage{algorithm}
\usepackage{algorithmic}
\usepackage{amsfonts}
\usepackage{amsmath}
\usepackage{subcaption}
\usepackage{xcolor}
\usepackage{booktabs}

\usepackage{newfloat}
\usepackage{listings}
\DeclareCaptionStyle{ruled}{labelfont=normalfont,labelsep=colon,strut=off} 
\floatstyle{ruled}
\newfloat{listing}{tb}{lst}{}
\floatname{listing}{Listing}
\nocopyright 

\title{
2-hop Neighbor Class Similarity (2NCS): A graph structural metric indicative of graph neural network performance
}
\author{
    Andrea Cavallo,\textsuperscript{\rm 1}\thanks{Work done during an internship in the Huawei Munich Research Center, AI4Sec team.}
    Claas Grohnfeldt,\textsuperscript{\rm 2}
    Michele Russo,\textsuperscript{\rm 2}
    Giulio Lovisotto,\textsuperscript{\rm 2}
    Luca Vassio\textsuperscript{\rm 1}
}
\affiliations{
    \textsuperscript{\rm 1}Politecnico di Torino, Turin, Italy\\
    \textsuperscript{\rm 2}Huawei Munich Research Center, Munich, Germany\\

    a.cavallo@studenti.polito.it, \{claas.grohnfeldt, michele.russo1, giulio.lovisotto\}@huawei.com,
    luca.vassio@polito.it

}

\begin{document}

\maketitle

\pubnote{Accepted at the 3\textsuperscript{rd} Workshop on Graphs and more Complex structures for Learning and Reasoning (GCLR) at AAAI 2023}

\begin{abstract}
 Graph Neural Networks (GNNs) achieve state-of-the-art performance on graph-structured data across numerous domains. Their underlying ability to represent nodes as summaries of their vicinities has proven effective for homophilous graphs in particular, in which same-type nodes tend to connect. On heterophilous graphs, in which different-type nodes are likely connected, GNNs perform less consistently, as neighborhood information might be less representative or even misleading. On the other hand, GNN performance is not inferior on all heterophilous graphs, and there is a lack of understanding of what other graph properties affect GNN performance.
 
 In this work, we highlight the limitations of the widely used \textit{homophily ratio} and the recent \textit{Cross-Class Neighborhood Similarity} (CCNS) metric in estimating GNN performance. To overcome these limitations, we introduce \textit{2-hop Neighbor Class Similarity} (2NCS), a new quantitative graph structural property that correlates with GNN performance more strongly and consistently than alternative metrics. 2NCS considers two-hop neighborhoods as a theoretically derived consequence of the two-step label propagation process governing GCN's training-inference process. Experiments on one synthetic and eight real-world graph datasets confirm consistent improvements over existing metrics in estimating the accuracy of GCN- and GAT-based architectures on the node classification task.
\end{abstract}

\section{Introduction}
Graph Neural Networks (GNNs) are becoming the \textit{de facto} default type of architecture to solve a variety of tasks on graph-structured data. GNNs generate a latent representation for each node in a graph based on information aggregated over the node's neighborhood including the node itself. Due to the underlying assumption that a node's neighborhood represents the node better than its features on their own, GNNs work particularly well on homophilous graphs, i.e., graphs in which nodes are more likely connected to nodes of the same class~\cite{zhu_beyond_2020, pei_geom-gcn_2019}. Moreover, recent studies prove that the reverse conclusion of GNNs performing worse on heterophilous graphs cannot be drawn in general~\cite{ma_is_2021}. Table~\ref{tab:std_GNN_results} shows that, for
some heterophilous graphs, a simple Multilayer Perceptron (MLP) applied to the node features outperforms GNNs on node classification, 
indicating that the graph structure, in those cases, is detrimental. However, on 
other heterophilous graphs, the best-performing GNN model largely improves over MLP. This motivates the need for deeper investigations of graph properties that affect GNN performance. 

In this work, we analyze the learning process of a simple yet representative GNN that does not consider node features. Based on this analysis, we derive a measurable structural graph property that - in contrast to other metrics related to GNN performance such as the standard \textit{homophily} and CCNS~\cite{ma_is_2021} - involves not only the 1-hop, but also the 2-hop neighborhood. We motivate and demonstrate that the label distribution of the second-hop neighborhood significantly influences the node classification capability of a GNN. More precisely, a node $u$ in a graph is more likely to be classified correctly if most of its neighbors have neighbors that are of the same class as $u$. To quantify this property, we introduce a new metric, \textit{2-hop Neighbor Class Similarity} (2NCS), and we perform an extensive evaluation to show that it correlates with GNN performance on real and synthetic graphs more strongly than related metrics. 

\begin{table*}
    \small
    \centering
    \begin{tabular}{ c c c c c c c c c }
    \hline
    & \verb+Cornell+ & \verb+Texas+ & \verb+Wisconsin+ & \verb+Film+ & \verb+Chameleon+ & \verb+Squirrel+ & \verb+Cora+ & \verb+Citeseer+ \\
    \(h\) & 0.30 & 0.11 & 0.21 & 0.22 & 0.23 & 0.22 & 0.81 & 0.74 \\
    \hline
    GCN & 50.27$\pm$7.57 & 57.03$\pm$5.05 & 50.98$\pm$4.88 & 23.27$\pm$0.94 & \textbf{67.43$\pm$1.95} & \textbf{50.49$\pm$1.54} & 84.29$\pm$0.98 & 73.25$\pm$1.42 \\
    GAT & 61.89$\pm$5.05 & 52.16$\pm$6.63 & 49.41$\pm$4.09 & 27.44$\pm$0.89 & 60.26$\pm$2.50 & 40.72$\pm$1.55 & \textbf{87.30$\pm$1.10} & \textbf{76.55$\pm$1.23} \\
    MLP & \textbf{81.89$\pm$6.40 }& \textbf{80.81$\pm$4.75} & \textbf{85.29$\pm$3.31} & \textbf{36.53$\pm$0.70} & 46.21$\pm$2.99 & 28.77$\pm$1.56 & 75.69$\pm$2.00 & 74.02$\pm$1.90 \\
    \hline
    \end{tabular}
    \caption{Average node classification accuracy and standard deviation over the ten different train/val/test splits from \cite{pei_geom-gcn_2019} for GCN, GAT and MLP on real-world datasets with varying \textit{edge homophily ratio} \(h\). Best results are in \textbf{bold}. Results for MLP and GAT are adopted from \cite{bodnar_neural_2022}, results for GCN are obtained from our own experiments.}   
    \label{tab:std_GNN_results}
\end{table*}

\section{Notation}

Let \(G = (V,E)\) be an unweighted and undirected graph, where \(V\) is the set of nodes and \(E\) is the set of edges. The connectivity information about nodes in \(G\) is represented by its adjacency matrix \(A\in \mathbb{R}^{n\times n}\), where \(n = |V|\) is the number of nodes and \(A_{uv}\) equals 1 if nodes \(u, v\in V\) are adjacent to each other, and 0 otherwise. Each node $u$ is associated with a feature vector \(x_u\) of size \(f\), and the complete set of features in the graph is denoted by \(X\in \mathbb{R}^{n\times f}\). Each node $u$ is also associated with a label \(y_u\in C\) representing the class of the node, where \(C\) is the set of classes. $V_c$ is the set of nodes belonging to class $c$. The set of adjacent nodes to a node \(u\) is called node \(u\)'s \textit{neighborhood} and is denoted by \(N(u)\). \(N'(u):=N(u)\cup\{u\}\) denotes the union of node $u$ and its neighbors. \(I\in \mathbb{R}^{n\times n}\) is the identity matrix.

\section{Motivation and related work}
The difficulties of using GNNs on non-homophilous graphs are pointed out by several works, which address the problem by introducing new GNN architectures to improve node representation capabilities~\cite{zhu_beyond_2020, jin_universal_2021, pei_geom-gcn_2019, yan_two_2021}.
Homophily is commonly quantified using the \textit{edge homophily ratio} $h$, which is defined as:
\begin{equation}
    h = \frac{|\{(u,v):(u,v)\in E \land y_u=y_v\}|}{|E|},
    \label{eq:edge_hom_rat}
\end{equation}
i.e. the fraction of edges connecting nodes with the same label.
Nevertheless, the \textit{edge homophily ratio} does not always capture GNN performance accurately.
To highlight this shortcoming, we perform a comparative evaluation of $h$ and node classification accuracy on the eight real-world graphs presented in the Appendix; Table~\ref{tab:std_GNN_results} reports the results.
Other than a standard MLP, we use two of the most popular GNNs, the Graph Convolutional Network (GCN)~\cite{kipf_semi_2017} and the Graph Attention Network (GAT)~\cite{velickovic_graph_2018}.
GCN aggregates information by averaging the transformed features of the neighbors and the target node, whereas in GAT the contributions of the neighbors are weighted by attention coefficients.
Table~\ref{tab:std_GNN_results} shows that high values of $h$ do not always correspond to good GNN performance: for GCN, examples such as \texttt{Chameleon} and \texttt{Squirrel} present relatively good accuracies ($\geq$50\%) while having low values of $h$ ($\leq$0.23).

Aware of this limitation, ~\cite{ma_is_2021} present a deeper analysis of the relationship between heterophily and GNN performance.
The authors prove that a GCN is more likely to perform well on a graph if nodes with the same label share similar distributions of labels in the neighborhoods and different classes have distinguishable patterns, regardless of the homophily.
To measure this property,~\cite{ma_is_2021} introduce \textit{Cross-Class Neighborhood Similarity} (CCNS), which is defined as:
\begin{equation}
    \label{eq:ccns}
    s(c,c') = \frac{1}{|V_c||V_{c'}|}\sum_{u\in V_c, v\in V_{c'}}\cos(d(u),d(v)),
\end{equation}
where $d(u)$ is the empirical label histogram over $|C|$ classes of node $u$'s neighbors and $\cos$ is the cosine similarity function.
Despite its effectiveness in the experiments reported in \cite{ma_is_2021}, CCNS has two limitations.
First, it relies on the assumption that node features and node labels are strongly correlated, which is not always observed on real-world graphs.
In fact, Table~\ref{tab:std_GNN_results} shows that MLP performs poorly on \verb+Chameleon+ and \verb+Squirrel+, thus indicating non-informative node features.
Secondly, CCNS does not correlate to the high GCN accuracy on the heterophilous graphs \verb+Chameleon+ and \verb+Squirrel+.
In the following we introduce a new metric to better estimate GNN performance.

\section{2NCS: 2-hop Neighbor Class Similarity}

Here, we analyze how node embeddings are generated in GNNs, focusing on a simplified GCN model. We choose to work on GCN due to its wide usage and simplicity, but we also use GAT for evaluation.
Based on this analysis, we introduce a novel metric for GNN performance named 2NCS.

\subsection{A simplified GCN model}
\label{sec:simplified_gcn}
One limitation of CCNS is the assumption that features of nodes belonging to the same class are drawn from the same distribution. Although this might seem a reasonable assumption in general, experiments on real-world graphs, such as \verb+Squirrel+, on which the graph-unaware MLP hardly improves over random guessing for node classification, refute its general validity. Moreover, as node features are not always available in practice~\cite{rossi_thenetwork_2015}, it is common to adapt GNNs to the absence of node features~\cite{Zhu_graph_2021}. 

This motivates the introduction of a simplified single-layer GCN, in which the feature matrix is equal to the identity matrix \(X=I\) and \(A\) is not row-normalized.
Formally, the simplified GCN model can be expressed as follows:
\begin{equation}
    \label{eq:simplified_gcn}
    H = \text{softmax}(\tilde{A}XW) = \text{softmax}(\tilde{A}IW) = \text{softmax}(\tilde{A}W),
\end{equation}
where \(H \in \mathbb{R}^{n\times |C|}\) are the class probabilities for each node, $\tilde{A}=A+I$ is the adjacency matrix with added self-loops and \(W \in \mathbb{R}^{n\times |C|}\) is a learnable weight matrix. These simplifications allow for a thorough analysis of the model's representation capabilities and the identification and characterization of structural graph information that impacts model performance. Note that the simplified GCN is similar to the LINK model introduced by \cite{zheleva_to_2009}.

\subsection{Learning process of the simplified GCN}
To understand which properties of the graph structure affect GCN performance, we analyze how node embeddings $H$ are generated by the simplified GCN, and how model weights $W$ are learned.
We consider standard gradient descent-based optimization where a suitable classification loss (e.g., cross-entropy $\mathcal{L}_{\text{CE}}$) is minimized on a set of training examples.

In the simplified GCN model shown in Eq.~\eqref{eq:simplified_gcn}, each row \(W_u\) can be interpreted as an embedding of \(u\) before aggregation over its neighbors, whereas \(H_u\) is the embedding of \(u\) after aggregation and softmax normalization, which corresponds to the class probabilities for the node.
Following Eq.~\eqref{eq:simplified_gcn}, \(H_u\) is computed by aggregating information from the $W_v$ embeddings corresponding to $u$'s neighbors and itself:
\begin{equation}
    \label{eq:4}
    H_u = \text{softmax}\left(\textstyle \sum_{v\in N'(u)}W_v\right)
\end{equation}

Assuming a batch size of one for simplicity, we can reason about which information influences a certain $W_v$ during back-propagation.
As per Eq.~\eqref{eq:4}, computing the loss at a node $z$ will trigger an update on a certain $W_v$ (i.e., $\nabla_{W_v}\neq 0$) if $v$ is a neighbor of $z$:
\begin{equation}\label{eq:6}
\nabla_{W_v}\mathcal{L}_{\text{CE}}(H_z, y_z) \neq \: 0 \: \hspace{2mm}\text{iff}\hspace{2mm}\: v \in N'(z).
\end{equation}
Since the training objective is to minimize cross-entropy, an embedding $W_v$ is updated such that the output probability corresponding to the observed label $y_z$ is increased and the probabilities of the non-observed labels are decreased.
Generalizing to the case of batch size greater than one, $W_v$ will learn higher values for the class observed more often during training, i.e., the most common class among the node $v$ itself and its neighbors $N(v)$. 

\begin{figure}[t]
\centering
\includegraphics[width=.5\linewidth]{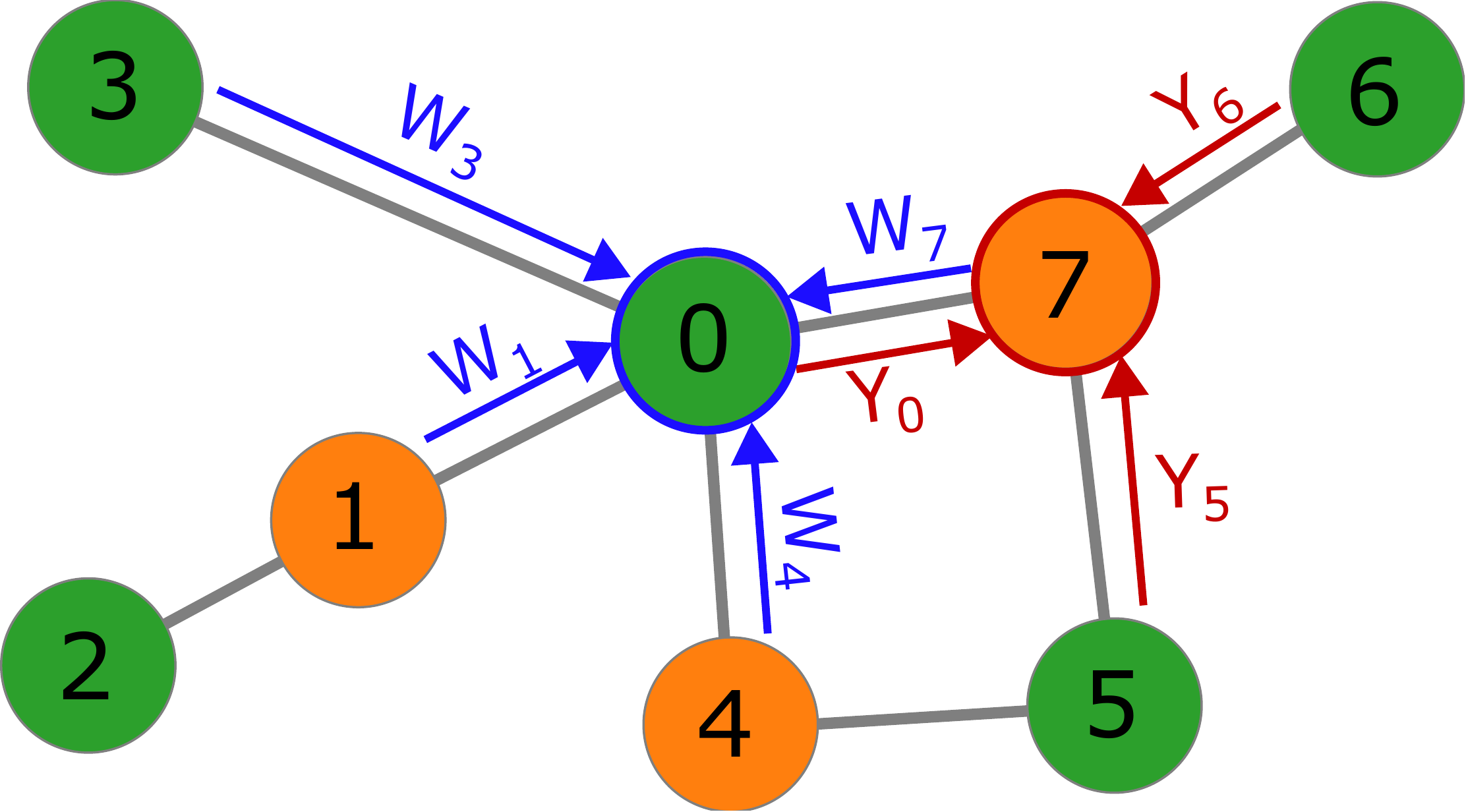}
\caption{Visualization of what information influences $H_0$ in the simplified GCN model. Different colors indicate different node labels. At training time, embeddings $W_v$ are updated based on the labels of their neighbors (red arrows). For example, $W_7$ is updated based on the label information propagating from nodes 0, 5, 6 and 7. At inference time, $H_0$ depends directly on the learned weights of its neighbors (blue arrows) and itself, i.e., $H_0=\text{softmax}(W_0+W_1+W_3+W_4+W_7)$, and  indirectly on the labels of its 2-hop neighbors due to the label propagation process. Self-loops are not shown for clarity.}
\label{fig:graph_2ncs_ex}
\end{figure}

Based on these considerations, we now analyze which elements in the graph affect the class probabilities $H_u$ of node $u$ at inference time, see also Figure~\ref{fig:graph_2ncs_ex} for reference.
As discussed above, each term \(W_v\) contributing to $H_u$ in Eq.~\eqref{eq:4} describes the class distribution of $v$'s neighbors as observed during training.
As a consequence, $H_u$ depends not only on $u$'s neighbors $N'(u)$, but also on the neighbors of its neighbors $z$, i.e. its 2-hop neighbors: $ z \in N'(v) \setminus \{u\} \: \forall v \in N'(u)$. 
Therefore, we can state that the simplified GCN model is more likely to classify a node $u$ correctly if \textit{the majority of $u$'s neighbors have, for the most part, neighbors with the same label as $u$.}

\subsection{2NCS: a graph structural metric}
Based on the analyses described above, we introduce \textit{2-hop Neighbor Class Similarity} (2NCS), a new local graph metric to quantify the probability of the simplified GCN classifying a node correctly. For a given node $u$, we define it as follows:
\begin{equation}
\small
    \label{eq:2ncs_single_node}
    \textnormal{2NCS}_u = \frac{1}{|N'(u)|}\sum_{v\in N'(u)}\frac{|\{z:z\in N'(v) \setminus \{u\} \land y_z = y_u\}|}{|N'(v)|-1}
\end{equation}
Note that node $u$ is removed from the count of neighbors with the same label since the goal is to understand whether its label can be correctly predicted given the rest of the graph structure. We can also interpret this problem as the training of a simplified GCN on the whole graph but node $u$ and the evaluation on node $u$ as test set. In this case, the label of node $u$ is not available during training; therefore, it would be inaccurate to include it in the 2NCS computation. The range of possible values of 2NCS is the interval $[0,1]$, where \(\textnormal{2NCS}_u=0\) means that \textit{no} 1- and \textit{no} 2-hop neighbor of node $u$ shares node $u$'s label, and \(\textnormal{2NCS}_u=1\) means that \textit{all} of node $u$'s 1- and 2-hop neighbors share node $u$'s label. 

We define the graph-level 2NCS as the average of $\textnormal{2NCS}_u$ for all nodes $u$ in the graph. Considering the property it represents, \textit{a simplified GCN is more likely to classify nodes with high 2NCS correctly, and the graph-level 2NCS is informative about the expected performance of a simplified GCN}.

\section{Analysis and evaluation}

In this section, we evaluate how informative 2NCS is about the suitability of the graph's topology and class distribution for a GNN to perform node classification. 

\begin{figure}[t]
\centering
\includegraphics[width=0.88\linewidth]{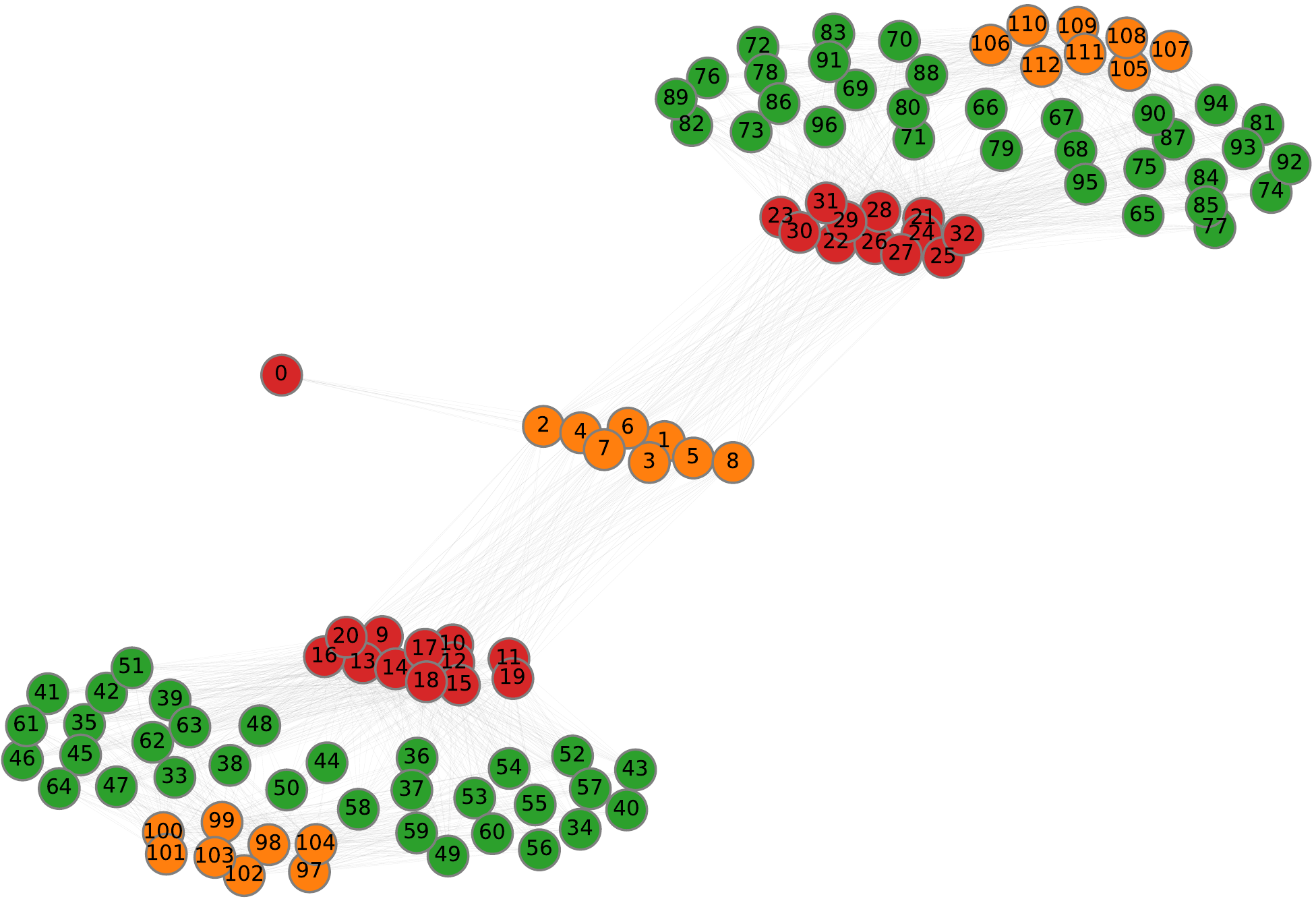}
\caption{Example of a graph where 2NCS is more informative than homophily and CCNS for GCN performance. 
Different colors indicate different node labels. The goal is to classify node 0.
Orange nodes 1-8 are densely connected to the three groups of red nodes 0, 9-20 and  21-32. The green group 33-64 is densely connected to red nodes 9-20 and orange nodes 97-104.  The green group 65-96 is densely connected to red nodes 21-32 and orange nodes 105-112. 
}
\label{fig:art_graph_2ncs_ex1}
\end{figure}

\subsection{2NCS on synthetic graphs}
We begin by demonstrating that there exist graphs on which 2NCS is more informative about GCN performance than related metrics, namely $h$ and CCNS. Figure~\ref{fig:art_graph_2ncs_ex1} depicts a graph that we created so that node 0's local homophily ratio equals zero and $\textnormal{CCNS}_0=0.24$.
These metrics indicate that the graph structure does not help a GCN classify node 0 correctly.
Note that, in this experiment, we ignore node features by letting $X=I$.
When we train a GCN on every node but node 0 and test the prediction accuracy on node 0, the node is classified correctly.
Remarkably, the high value of 2NCS ($\textnormal{2NCS}_{\text{0}}=0.85$) for node 0 explains this behavior correctly. Indeed, a GCN correctly classifies node 0 because it is connected to nodes 1-8, which are connected to nodes of the same class as node 0 (red). This pattern allows for high node classification performance of GCNs, but it is not captured by the other two metrics.

\subsection{2NCS on real-world graphs}


    
    

\begin{figure}[t]
\centering
    \addtolength{\leftskip} {-1cm}
    \addtolength{\rightskip}{-1cm}

\begin{subfigure}{0.25\textwidth}
    \includegraphics[width=\textwidth]{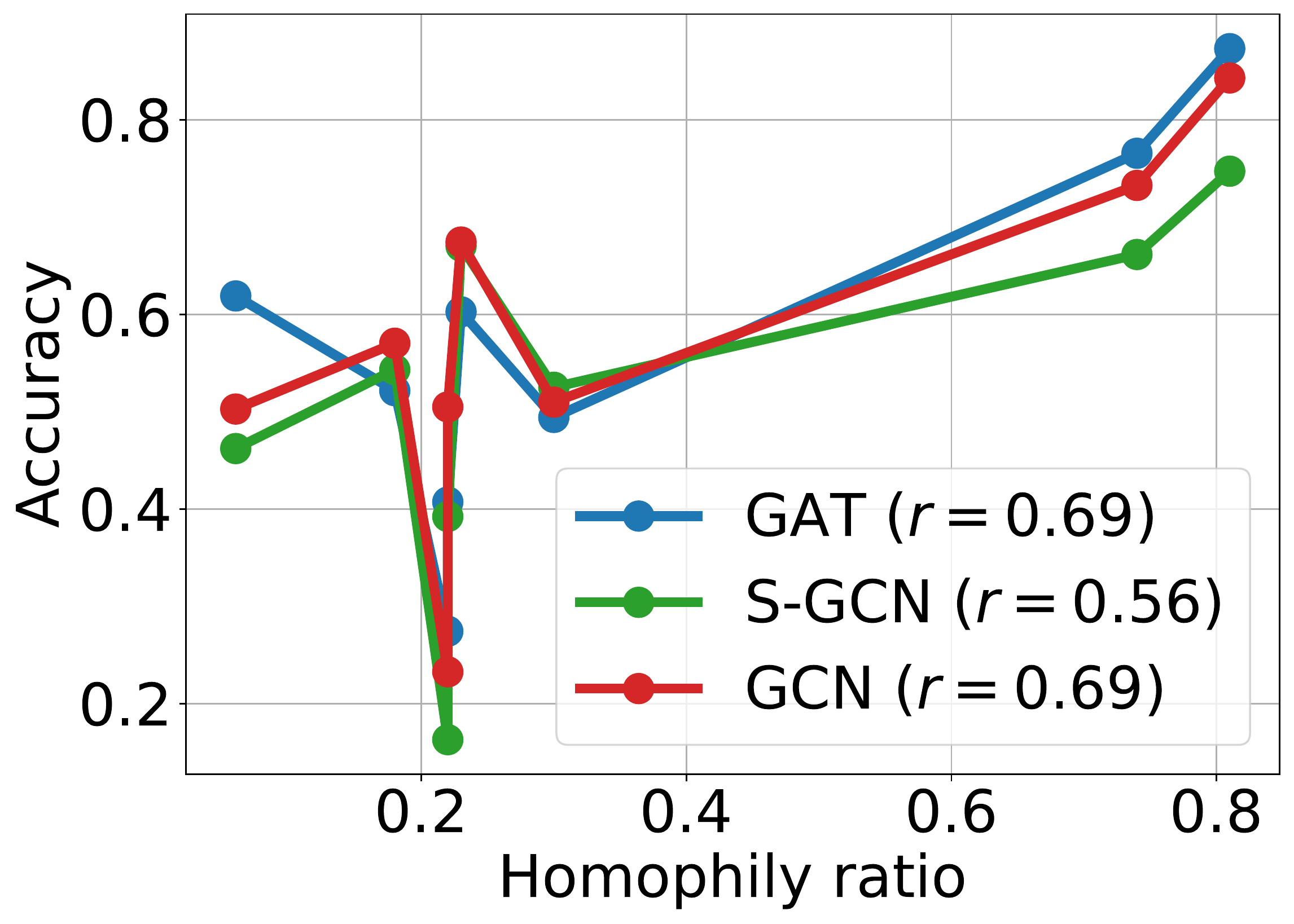}
\end{subfigure}
\begin{subfigure}{0.25\textwidth}
    \includegraphics[width=\textwidth]{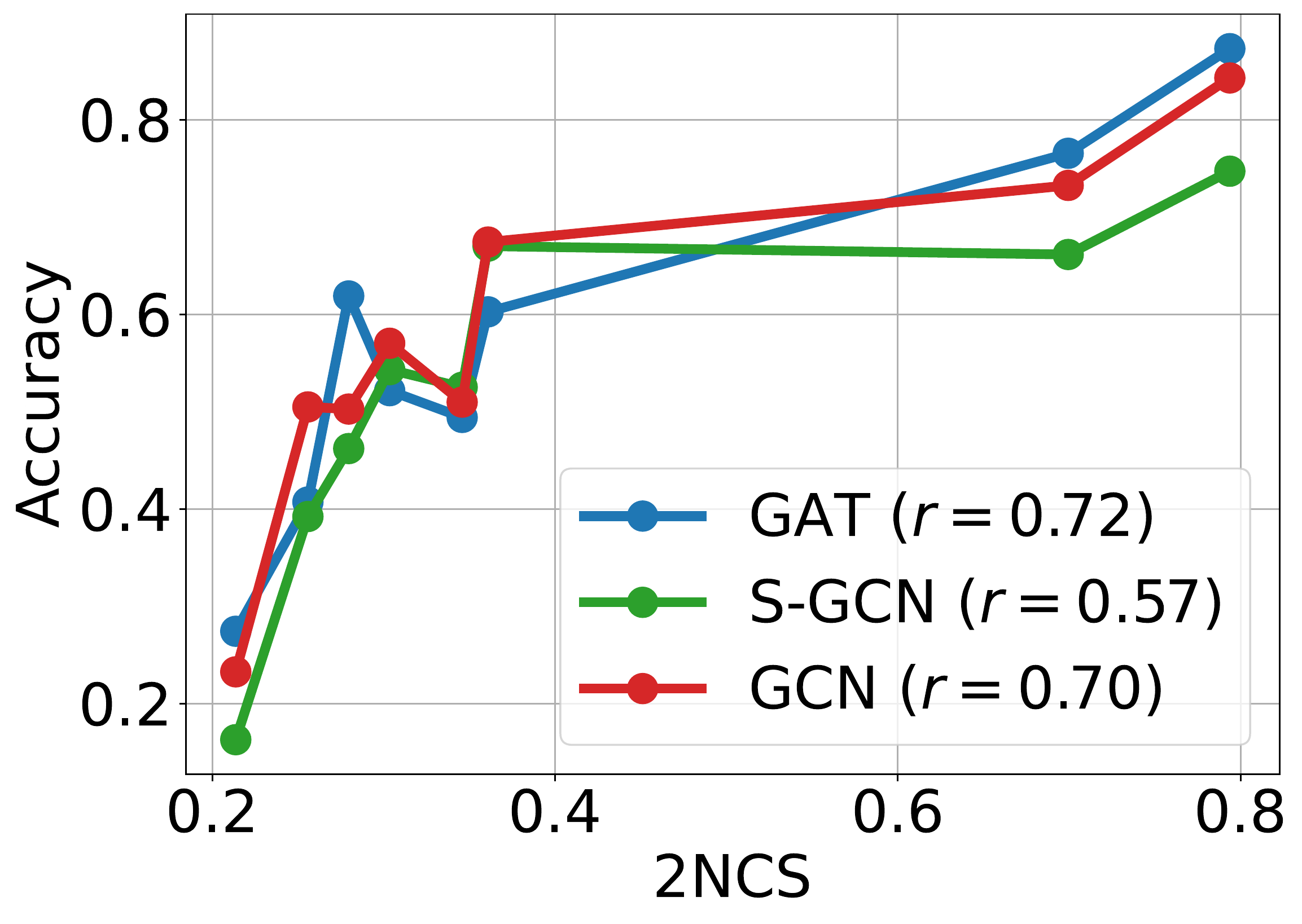}
\end{subfigure}

\caption{GNN performance versus homophily ratio and 2NCS on eight real-world graphs (each graph is a point in the data series). S-GCN is the simplified GCN model introduced in this work. $r$ is the sample Pearson correlation coefficient between the metric value and the model accuracy.}
\label{fig:2ncs_real_graph}
\end{figure}

\begin{figure}[t]
\centering
    \addtolength{\leftskip} {-1cm}
    \addtolength{\rightskip}{-1cm}
\begin{subfigure}{0.25\textwidth}
    \includegraphics[width=\textwidth]{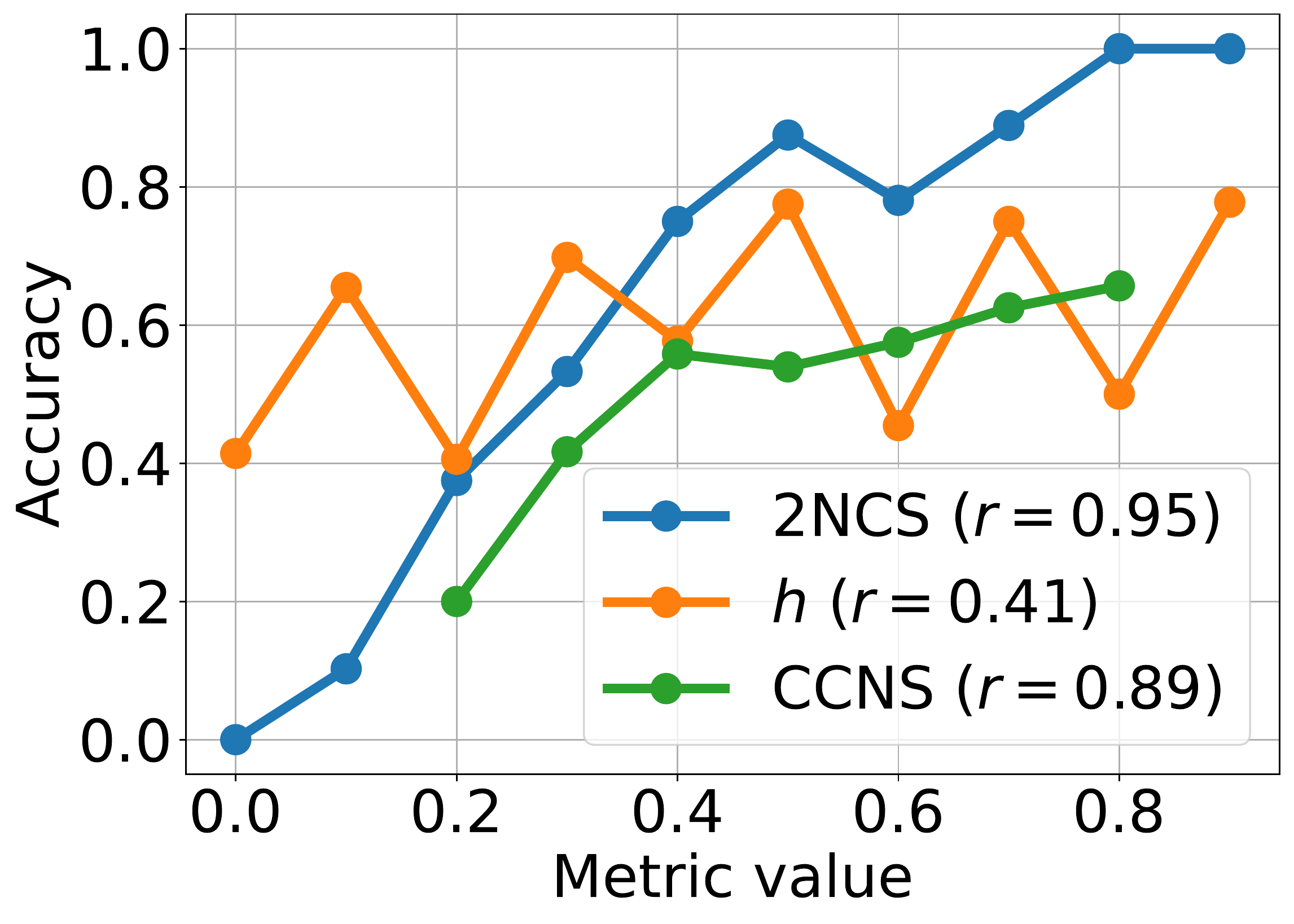}
    \caption{\texttt{Chameleon}, GCN}
\end{subfigure}
\begin{subfigure}{0.25\textwidth}
    \includegraphics[width=\textwidth]{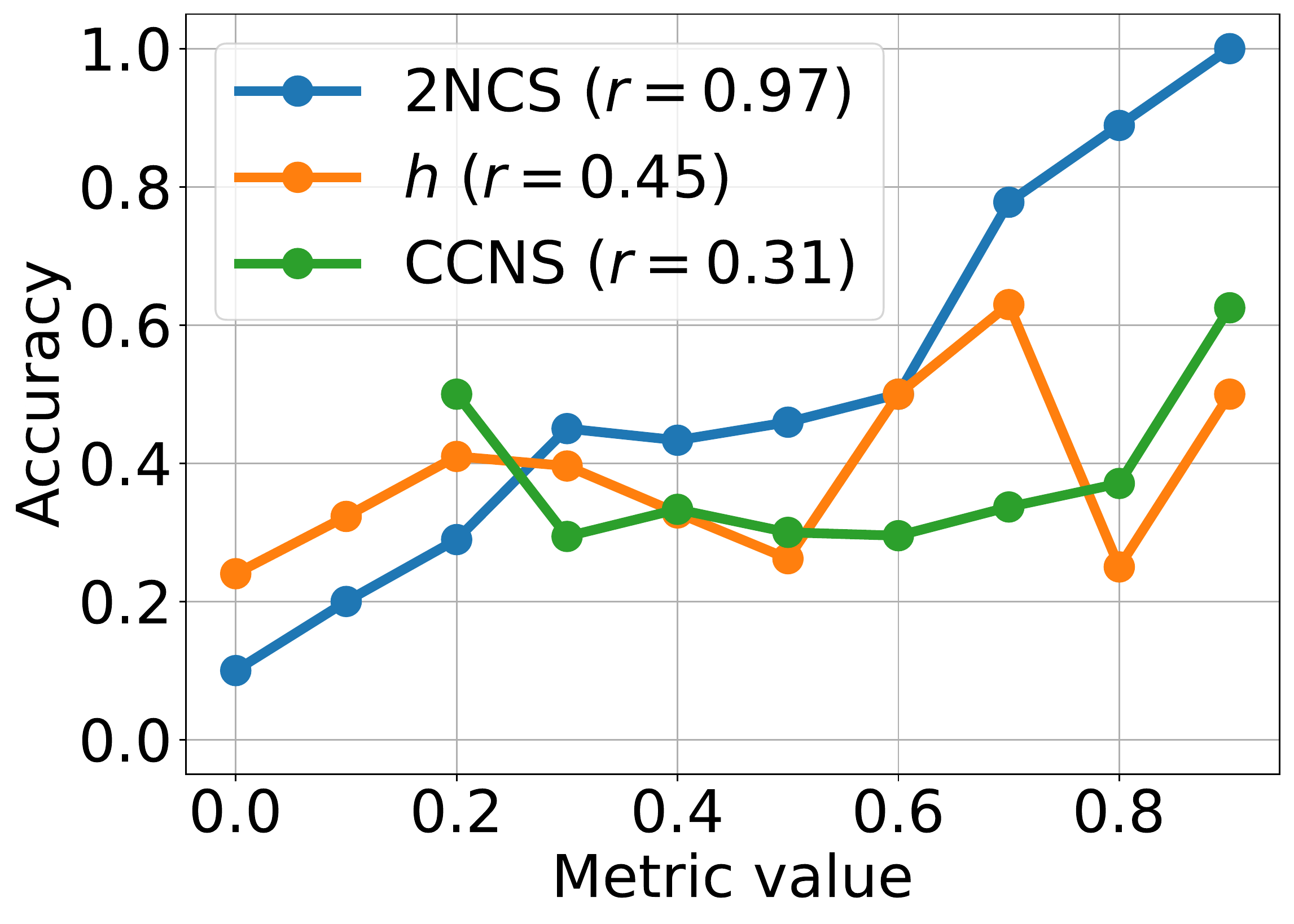}
    \caption{\texttt{Squirrel}, GCN}
\end{subfigure}
\begin{subfigure}{0.25\textwidth}
    \includegraphics[width=\textwidth]{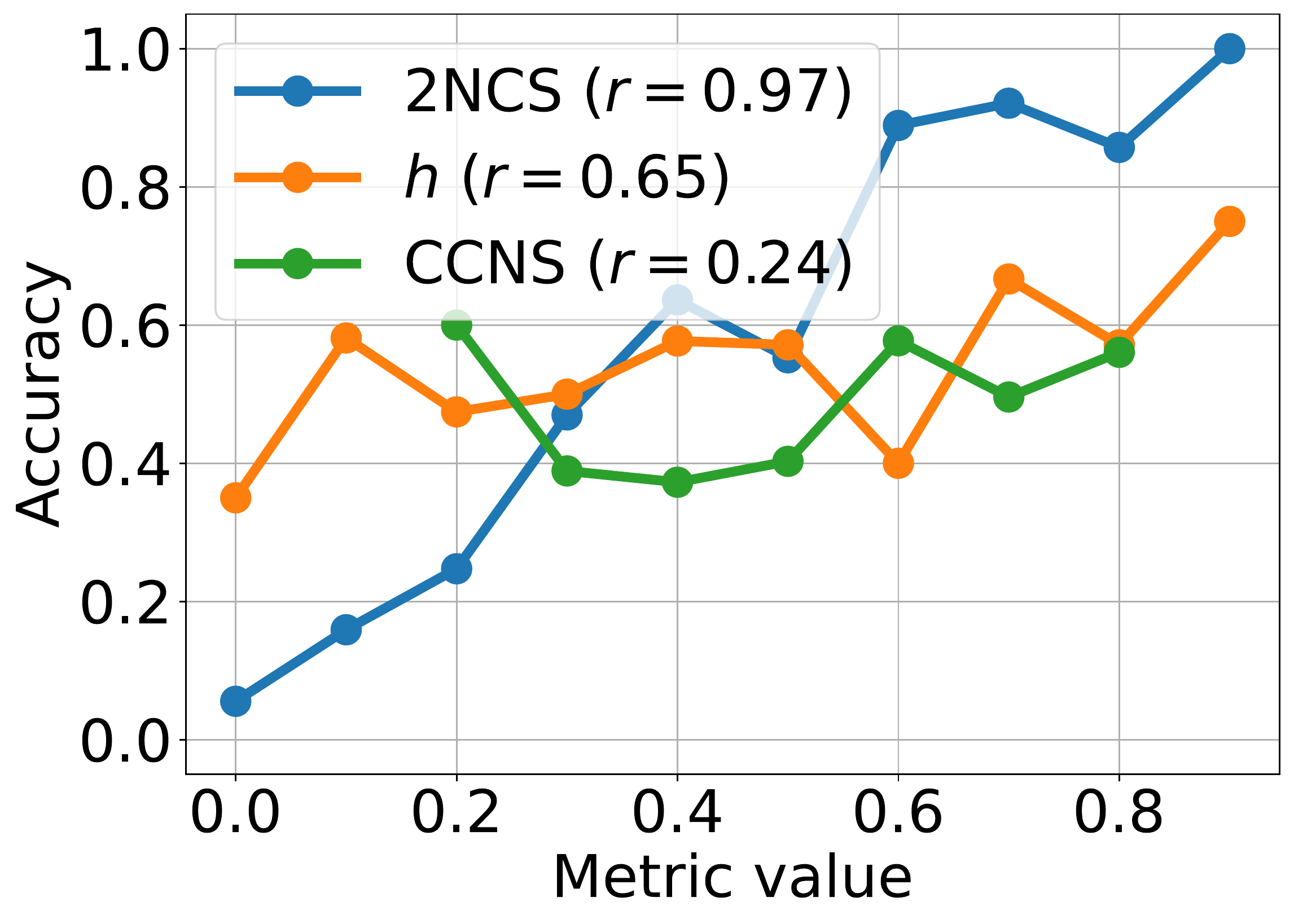}
    \caption{\texttt{Chameleon}, GAT}
\end{subfigure}
\begin{subfigure}{0.25\textwidth}
    \includegraphics[width=\textwidth]{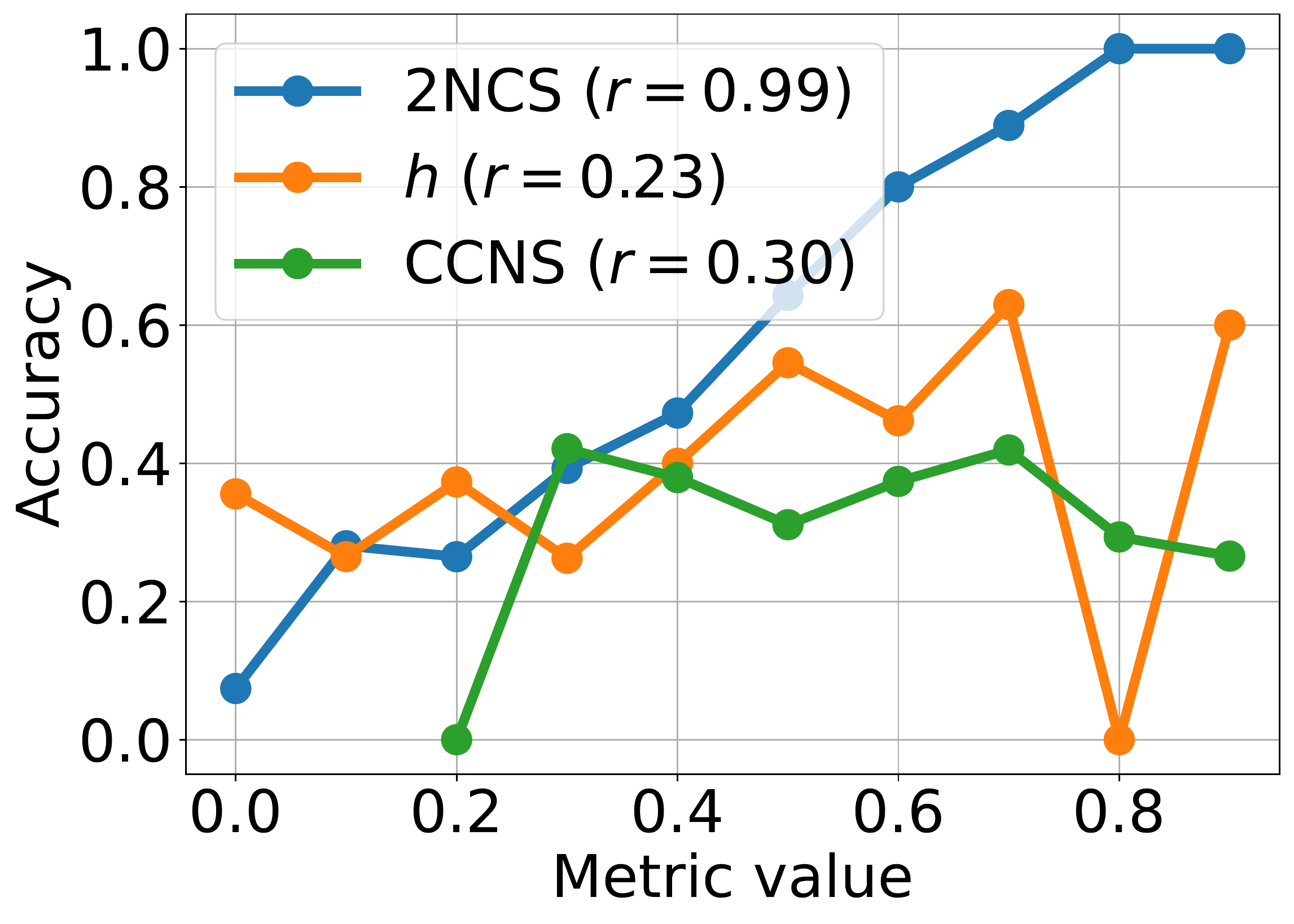}
    \caption{\texttt{Squirrel}, GAT}
\end{subfigure}
\caption{GCN and GAT accuracy versus node-level homophily ratio $h$, CCNS  and 2NCS on \texttt{Chameleon} and \texttt{Squirrel}. $r$ is the sample Pearson correlation coefficient between the metric value and the model accuracy.}
\label{fig:node_level_2ncs}
\end{figure}

\subsubsection{Graph-level 2NCS}


Figure~\ref{fig:2ncs_real_graph} depicts a performance comparison of different GNN architectures, namely GAT, GCN, and simplified GCN, over the homophily ratios (left) and 2NCS values (right) of the eight real-world datasets listed in Table~\ref{tab:std_GNN_results}.
The figure shows that there is no clear correlation between GNN performance and homophily ratio $h$: the lowest GNN accuracy does not correspond to the dataset with the lowest value of $h$. The right-hand-side plot in~Figure~\ref{fig:2ncs_real_graph} shows that the correlation between GNN performance and 2NCS is stronger.
We also notice that the accuracy values of the simplified GCN (introduced above) and the standard GCN are comparable, and that both align well with the 2NCS metric.

\subsubsection{Node-level 2NCS}
Here, we compare node-level 2NCS values with GCN and GAT classification accuracy.
As discussed above, we claim that nodes with higher 2NCS value will more likely be classified correctly by a GNN.
In order to validate this claim experimentally on real data, we plot the GCN and GAT classification accuracy against local, i.e. node-level, values of 2NCS, $h$, and CCNS, in Figure~\ref{fig:node_level_2ncs} for the \texttt{Chameleon} and \texttt{Squirrel} datasets. We observe that 2NCS has a noticeably higher correlation to both GCN and GAT accuracy than $h$ and CCNS. This is also confirmed by the values of the sample Pearson correlation coefficient $r$ between the different metrics and the classification accuracy.


\begin{figure}[t]
\centering
    \addtolength{\leftskip} {-1cm}
    \addtolength{\rightskip}{-1cm}

\begin{subfigure}{0.25\textwidth}
    \includegraphics[width=\textwidth]{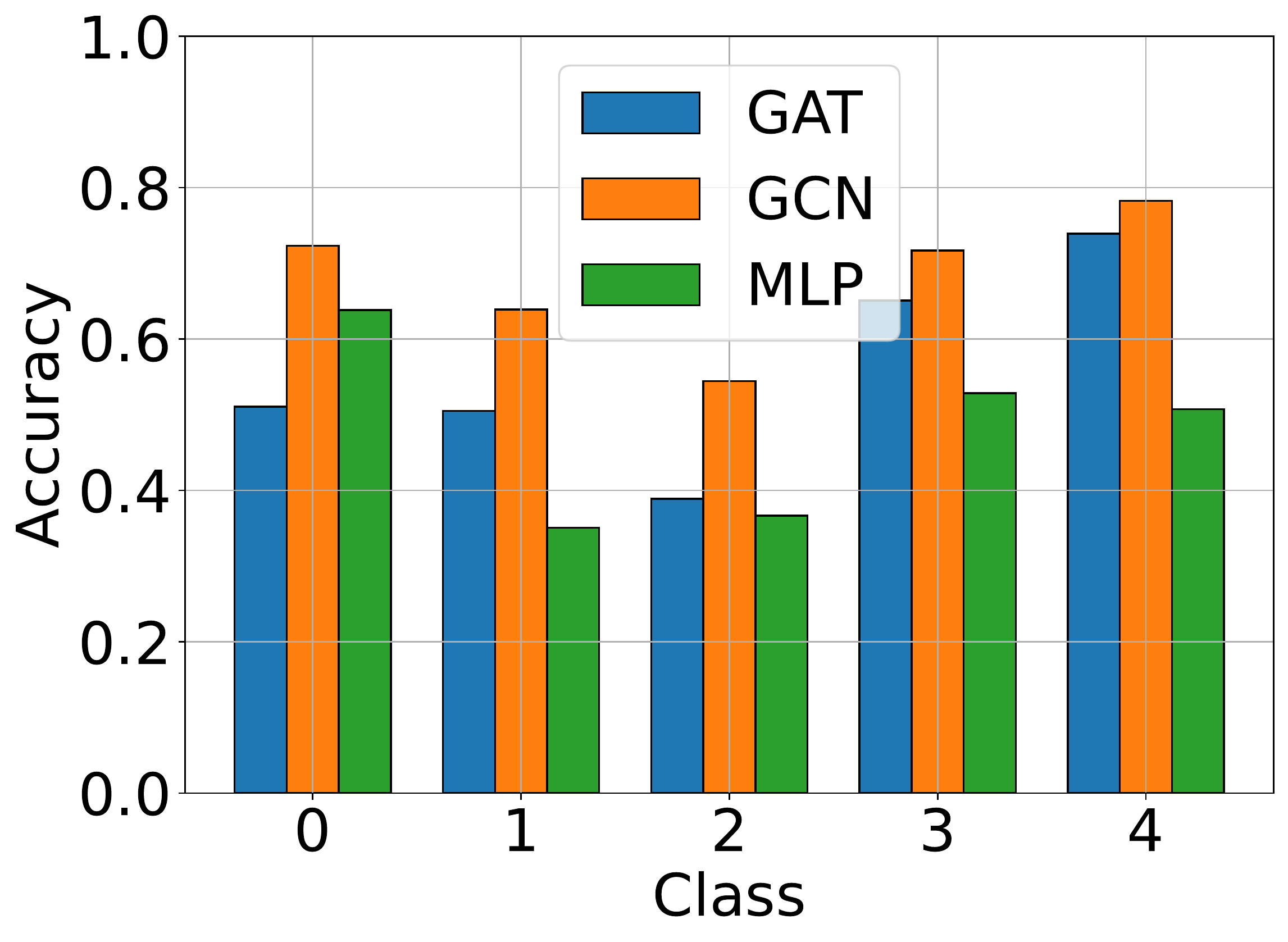}
    \caption{Chameleon}
    \label{fig:class_level_2ncs:cham}
\end{subfigure}
\begin{subfigure}{0.25\textwidth}
    \includegraphics[width=\textwidth]{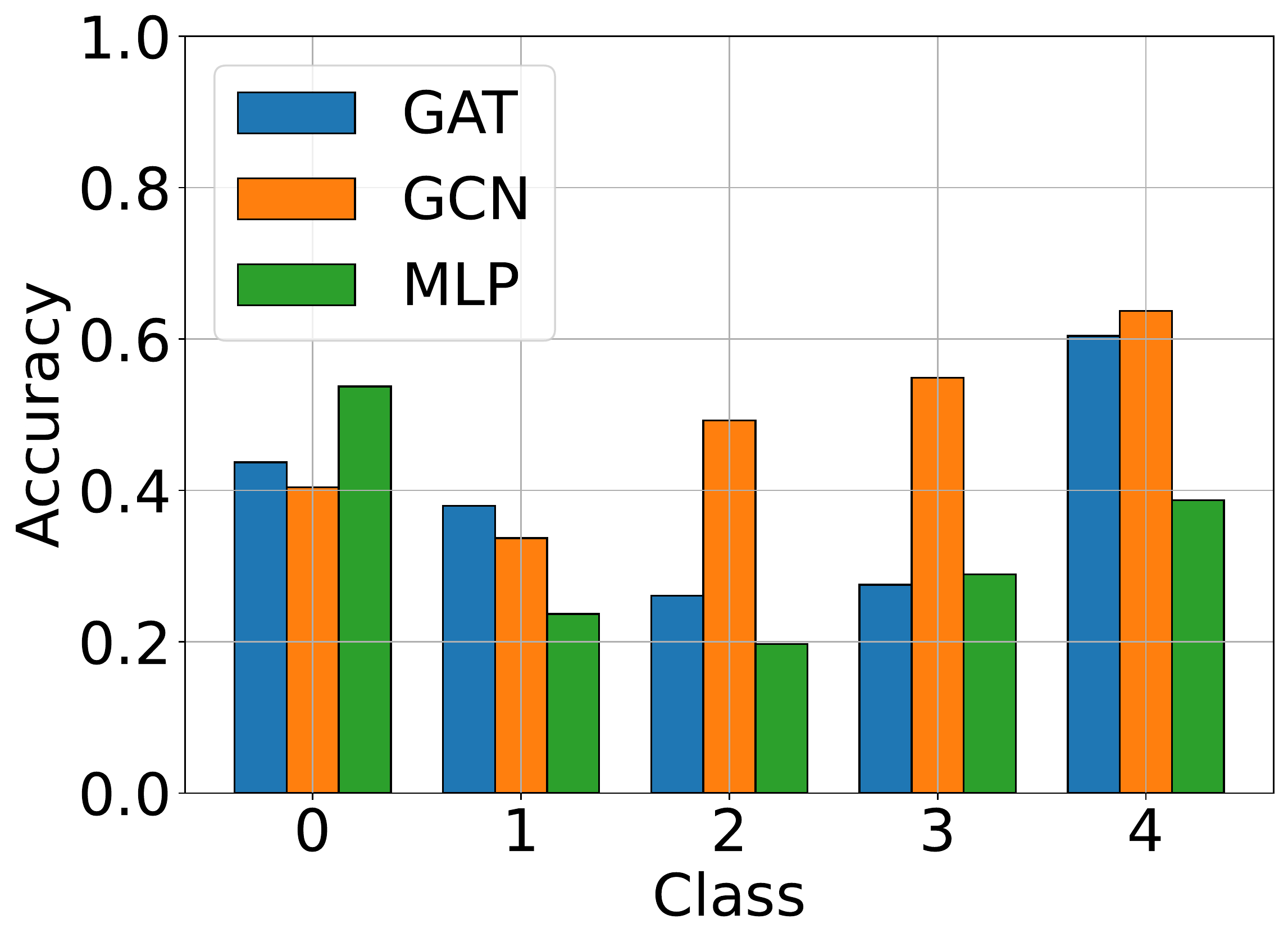}
    \caption{Squirrel}
    \label{fig:class_level_2ncs:squirrel}
\end{subfigure}
\caption{Per-class accuracy for MLP, GCN and GAT on \texttt{Chameleon} and \texttt{Squirrel}.}
\label{fig:class_level_2ncs}
\end{figure}

\begin{figure}[t]
\centering

\includegraphics[width=0.25\textwidth]{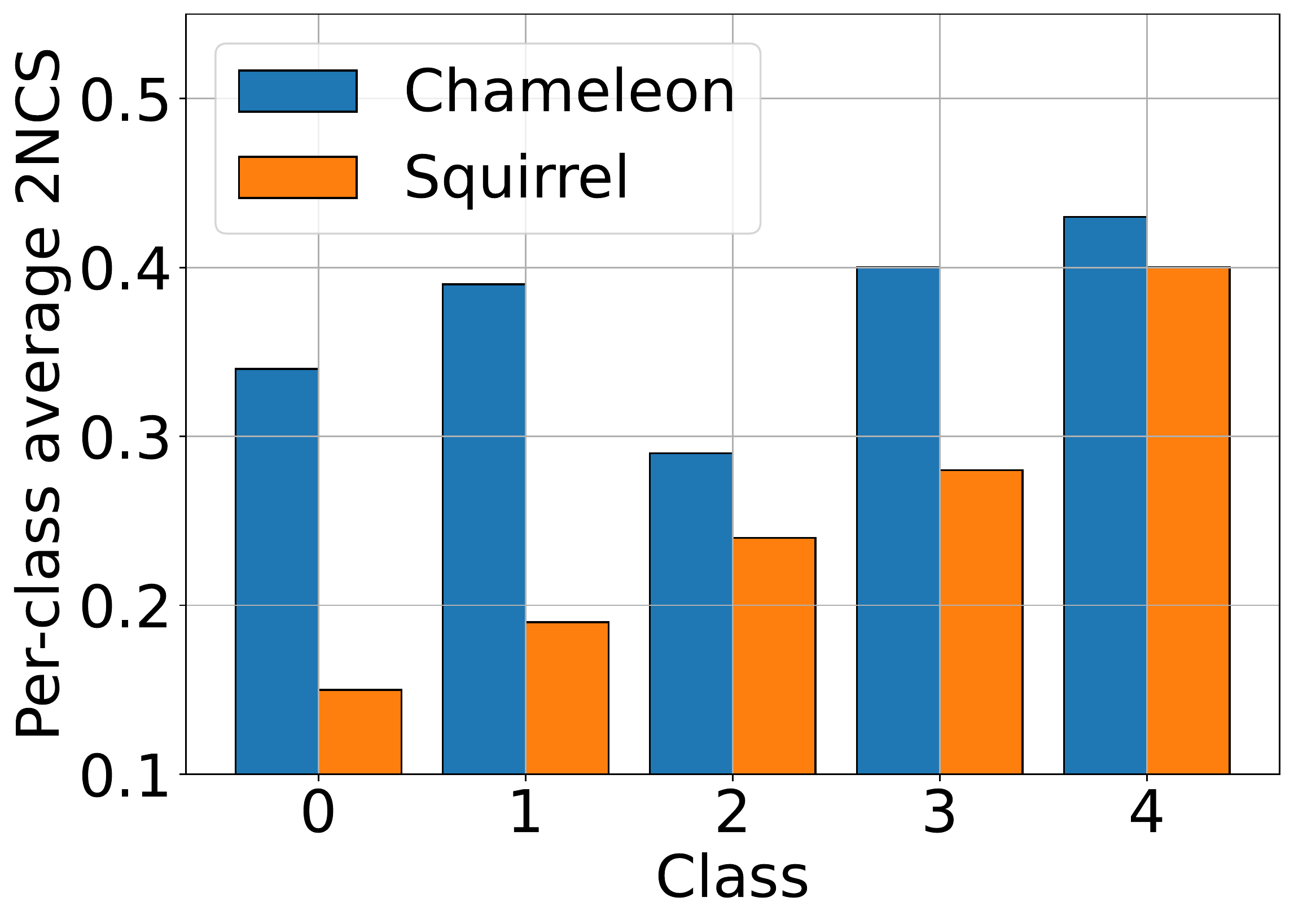}
\caption{Per-class 2NCS values for \texttt{Chameleon} and \texttt{Squirrel}.}
\label{fig:per_class_2ncs_values}
\end{figure}

\subsubsection{Per-class 2NCS}

Next, we investigate how 2NCS captures graph properties when we partition nodes by class.
We report the per-class 2NCS, i.e. average 2NCS over nodes belonging to that class, in Figure~\ref{fig:per_class_2ncs_values} for the two datasets \verb+Chameleon+ and \verb+Squirrel+.
Using MLP as a baseline for how informative node features are for node classification, we compare the accuracy of MLP, GCN and GAT for nodes in different classes on the datasets, see Figure \ref{fig:class_level_2ncs}.

The analysis presented above suggests that classes with higher 2NCS values will benefit more from the graph structure for classification.
This intuition is confirmed when we compare the values of per-class 2NCS reported in Figure~\ref{fig:per_class_2ncs_values} and the gaps between MLP, GCN and GAT performance reported in Figure~\ref{fig:class_level_2ncs}.
For example, on \verb+Chameleon+, the significant improvements of GCN and GAT over MLP for classes 1,3 and 4 match the high values of 2NCS for those classes (0.39, 0.40 and 0.43).
Similarly, on \verb+Squirrel+, Figure~\ref{fig:class_level_2ncs:squirrel} shows how, for class 0, MLP outperforms GCN and GAT, which is consistent with the low value of 2NCS found for that class (0.15), whereas for class 4 the large improvement of GCN and GAT over MLP corresponds to a high value of 2NCS (0.40).

\section{Conclusions}
\label{sec:2ncs_limitations}

We introduced 2NCS, a new structural metric to quantify the suitability of a graph for applications of graph neural networks, based on intuitions derived from a simplified GCN model.
Through experiments on real and synthetic graphs using both GCN and GAT, we validated that 2NCS is more indicative of GNN performance than the related metrics, homophily ratio and CCNS.

2NCS also has limitations. Its main weakness is that it neglects the impact of node features, since it only considers the label distribution in the graph. This limitation becomes detrimental in graphs where features are significantly more important than the structure. 
Furthermore, 2NCS is derived from reasoning on a \textit{1-layer} GCN, which is chosen due to its simplicity that allows for a thorough analysis of the learning-inference process. 
We leave a theoretically sound generalization to $k$-NCS, with $k>2$, and multi-layer GNNs as future work.
Finally, while the experimental evidence provided in this work confirms the theoretical derivation of high values of 2NCS consistently corresponding to high GNN performance, the opposite cannot be proved in general.




\bibliography{aaai23}

\appendix
\section{Datasets}
\label{sec:dataset}
The 8 real-world datasets used in this paper are commonly used in many works dealing with heterophilous graphs. Table \ref{tab:real_data_stat} shows their statistics. \verb+Cora+ and \verb+Citeseer+ are representative of homophilous settings, since their homophily ratio is high, whereas all the others are strongly heterophilous.
\begin{table*}
    \centering
    \footnotesize
    \addtolength{\leftskip} {-2cm}
    \addtolength{\rightskip}{-2cm}
    \begin{tabular}{ c c c c c c c c c }
    \toprule
    \textbf{Dataset} & \verb+Cornell+ & \verb+Texas+ & \verb+Wisconsin+ & \verb+Film+ & \verb+Chameleon+ & \verb+Squirrel+ & \verb+Cora+ & \verb+Citeseer+ \\
    \midrule
    \textbf{\#Nodes} & 183 & 183 & 251 & 7,600 & 2,277 & 5,201 & 2,708 & 3,327 \\
    \textbf{\#Edges} & 280 & 295 & 466 & 26,752 & 31,421 & 198,493 & 1,433 & 3,703 \\
    \textbf{\#Classes} & 5 & 5 & 5 & 5 & 5 & 5 & 7 & 6 \\
    \textbf{\#Features} & 1,703 & 1,703 & 1,703 & 931 & 2,325 & 2,089 & 1,433 & 3,703 \\
    \textbf{Homophily \(h\)} & 0.30 & 0.11 & 0.21 & 0.22 & 0.23 & 0.22 & 0.81 & 0.74 \\
    \textbf{CCNS} & 0.54 & 0.60 & 0.58 & 0.50 & 0.61 & 0.69 & 0.80 & 0.63 \\
    \textbf{2NCS} & 0.35 & 0.28 & 0.30 & 0.21 & 0.36 & 0.26 & 0.79 & 0.70 \\
    \bottomrule

    \end{tabular}
    \caption[Statistics of real-world datasets.]{Statistics of real-world datasets.}
    \label{tab:real_data_stat}
\end{table*}
\begin{table*}[h]
    \centering
    \small
    \addtolength{\leftskip} {-1cm}
    \addtolength{\rightskip}{-1cm}
    \begin{tabular}{ c c c c c }
    \toprule
    \textbf{Dataset} & \textbf{Batch} & \textbf{\#Epochs} & \textbf{Hidden} & \textbf{Dropout} \\
     & \textbf{size} &  & \textbf{size} & \textbf{rate} \\
    \midrule
    \verb+Cornell+, \verb+Texas+, \verb+Wisconsin+ & \{50,$|V|$\} & \{100,200,300\} & \{16,32\} & \{0.0,0.25,0.5\} \\
    \verb+Film+ & \{300\} & \{50\} & \{16,32\} & \{0.0,0.25,0.5\} \\
    \verb+Chameleon+, \verb+Squirrel+ & \{300,$|V|$\} & \{500,1000\} & \{16,32\} & \{0.0,0.25,0.5\} \\
    \verb+Cora+, \verb+Citeseer+ & \{300,$|V|$\} & \{100\} & \{16,32\} & \{0.0,0.25,0.5\} \\
    \bottomrule
    
    \end{tabular}
    \caption{Hyperparameters for GCN.} 
    \label{tab:gcnh_real_main_hyp}
\end{table*}All datasets are taken from the public code of \cite{pei_geom-gcn_2019}, with the exception of \verb+Chameleon+ and \verb+Squirrel+ that are taken from the public code of \cite{maurya_simplifying_2022}. We now present more details about each dataset. 

\begin{itemize}
    \item \textbf{Texas, Wisconsin and Cornell} are webpage datasets collected from the computer science departments of different universities by Carnegie Mellon University within the WebKB project\footnote{http://www.cs.cmu.edu/afs/cs.cmu.edu/project/theo-11/www/wwkb/}. Nodes represent web pages and edges are hyperlinks between them. Node features are bag-of-words representations of the web pages, which are manually classified into five categories: student, project, course, staff, and faculty. 
    \item \textbf{Film}, also referred to as \textbf{Actor}, is the actor-only induced subgraph of the film-director-actor-writer network \cite{tang_social_2009}. Nodes correspond to Wikipedia pages of actors and edges denote the co-occurrence of two actors on the same page. Node features are some keywords in the Wikipedia pages and labels are assigned by \cite{pei_geom-gcn_2019} based on words of the actors' Wikipedia pages. 
    \item \textbf{Chameleon and Squirrel} are Wikipedia pages on the specific topics of chameleons and squirrels. They were collected by \cite{rozemberczki_multi-scale_2021} and pre-processed by \cite{pei_geom-gcn_2019}. Nodes are Wikipedia pages and edges are mutual links between them. Node features indicate the presence of informative nouns on Wikipedia pages. Nodes are classified into five categories based on the average monthly traffic on the web page.
    \item \textbf{Cora and Citeseer} are standard citation networks where nodes represent papers and edges represent citations of one paper by another \cite{sen_collective_2008, namata_query_2012}. Node features are bag-of-words representations of papers and labels are the academic topics of the papers. These datasets are treated as undirected.
\end{itemize}

\section{Experimental details}

The results for GCN in Table \ref{tab:std_GNN_results} and GCN and simplified GCN in Figure \ref{fig:2ncs_real_graph} are obtained by performing a hyperparameter search among the values reported in Table \ref{tab:gcnh_real_main_hyp}. For GCN results in Figures \ref{fig:node_level_2ncs} and \ref{fig:class_level_2ncs}, the hyperparameters are the best ones from previous experiments. The results for GAT in Figures \ref{fig:node_level_2ncs} and \ref{fig:class_level_2ncs} are obtained using a model with 1 layer and 1 attention head trained for 100 epochs on \verb+Chameleon+ and 50 epochs on \verb+Squirrel+. The plots show results for nodes in the test set for one of the splits taken from \cite{pei_geom-gcn_2019}. 2NCS is computed only on the nodes of the training set, since labels of nodes in the evaluation and test set are supposed to be unknown during training. \\

\end{document}